\documentclass{article}
\usepackage{spconf,amsmath,graphicx}

\usepackage{xcolor}
\usepackage{mathtools}
\usepackage{amsfonts}
\usepackage{enumitem}
\usepackage{subcaption}
\usepackage{amsfonts}
\usepackage{url}
\usepackage{spconf}

\title{TODM: Train Once Deploy Many\\
Efficient Supernet-Based RNN-T Compression for on-device ASR models}

\name{ 
{Yuan Shangguan}, 
{Haichuan Yang},
{Danni Li\sthanks{Work done during AI Residency at Meta}},
{Chunyang Wu},
{Yassir Fathullah\sthanks{Work done during internship at Meta, University of Cambridge, UK}},
{Dilin Wang}}
\secondlinename{Ayushi Dalmia, Raghuraman Krishnamoorthi, Ozlem Kalinli}
\thirdlinename{Junteng Jia, Jay Mahadeokar, Xin Lei, Mike Seltzer, Vikas Chandra}

\address{Meta AI}

\begin{document}
\ninept

\maketitle

\begin{abstract}
Automatic Speech Recognition (ASR) models need to be optimized for specific hardware before they can be deployed on devices. This can be done by tuning the model's hyperparameters or exploring variations in its architecture. Re-training and re-validating models after making these changes can be a resource-intensive task.
This paper presents TODM (Train Once Deploy Many), a new approach to efficiently train many sizes of hardware-friendly on-device ASR models with comparable GPU-hours to that of a single training job. TODM leverages insights from prior work on Supernet, where Recurrent Neural Network Transducer (RNN-T) models share weights within a Supernet. It reduces layer sizes and widths of the Supernet to obtain subnetworks, making them smaller models suitable for all hardware types. We introduce a novel combination of three techniques to improve the outcomes of the TODM Supernet: adaptive dropout, an in-place Alpha-divergence knowledge distillation, and the use of ScaledAdam optimizer. We validate our approach by comparing Supernet-trained versus individually tuned Multi-Head State Space Model (MH-SSM) RNN-T using LibriSpeech. Results demonstrate that our TODM Supernet either matches or surpasses the performance of manually tuned models by up to a relative of 3\% better in word error rate (WER), while efficiently keeping the cost of training many models at a small constant.

\end{abstract}\vspace{-0.15cm}
\begin{keywords}
Supernet, RNN-T, on-device, efficiency, compression, knowledge distillations
\end{keywords}
%
\section{Introduction}
\label{intro}
End-to-end (E2E) all-neural automatic speech recognition (ASR) has gained attention for its compatibility with edge devices~\cite{graves2014towards, sainath2020streaming,shangguan2021dissecting}. Recurrent neural network transducer (RNN-T) is one of the most popular architectures for on-device E2E ASR~\cite{graves2012sequence, graves2013speech, li2021better, prabhavalkar2023endtoend}.
Modern applications of RNN-T ASR run on a variety of hardware, including central processing units (CPUs) on phones, tensor processing units (TPUs)~\cite{ding2022unified}, and other neural accelerators.
The process of optimizing on-device ASR models efficiently for each distinct hardware configuration necessitates fine-tuning model training hyperparameters and streamlining the model's architecture~\cite{shangguan2019optimizing}. This demands a substantial amount of computational resources, primarily in the form of GPU-hours or GPU cluster energy expenses in MegaWatt-hour~(MWh). Researchers must balance accuracy and size while efficiently managing training resources.

A potential approach is to leverage the Supernet~\cite{cai2019once,yu2019universally,yu2020bignas, yu2018slimmable, wang2021alphanet}. 
A Supernet is a weight-sharing neural network graph that can generate smaller subnetworks tailored for specific applications.
Prior work has attempted to facilitate the simultaneous training of a fixed number of models by implementing weight sharing capabilities.
Examples include work on RNN-T cascaded encoders~\cite{ding2022unified, narayanan2021cascaded}. 
These efforts, however, typically involve a limited number of networks, often around three, that only share portions of the RNN-T encoders.
Their primary objective is to streamline management for a few models, while utilizing distinct model decoders to improve model performances. Some also combine non-causal and causal encoders to enhance model accuracy. The training of such cascaded encoder systems consumes much more resources compared to a single model-training task, and it can not scale to a large number of different sizes of models.
Besides cascaded encoders, our previous work explored the concept of Supernet in the Omni-Sparsity DNN framework~\cite{yang2022omni}, in which a large number of ASR subnetworks could be derived from one Supernet training job. Furthermore, there have been efforts to expand the use of Omni-Sparsity DNN to train ASR models while accommodating various latency constraints for the subnetworks. This was demonstrated by training a non-streaming dense Supernet with sparse streaming subnetworks in~\cite{liu2023learning}.
Omni-Sparsity DNNs' reliance on structured sparsity, however, unnecessarily limits their hardware compatibility.



This paper presents Train Once Deploy Many (TODM), an approach to efficiently train many hardware-friendly on-device ASR models with comparable GPU-hours to a single training job. 
TODM leverages insights from prior work on Omni-sparsity DNN Supernet, but does not rely on sparsity. Instead, it drops layers and reduces layer widths in the Supernet to obtain optimized subnetworks for all hardware types.
We explain TODM Supernet in Section~\ref{sec:rnnt}. 
We improve its training outcome by introducing adaptive dropout, an in-place logit-sampled Alpha-divergence knowledge distillation mechanism, and ScaledAdam optimizer (Section~\ref{sec:trainsupernet}). 
We validate our approach by generating Supernet-trained sub-models using Evolutionary Search (described in Section~\ref{sec:evolutionarysearch}) on the validation set and comparing their performances to individually optimized ASR models using LibriSpeech~\cite{librispeechdataset} in Section~\ref{sec:exp}.
We discuss the results and why Supernet can function efficiently in Section~\ref{sec:results}.
This work is the first to use Supernet training to efficiently produce a variety of on-device ASR models. With similar training resources as one model, we can deploy many of different sizes, each with the optimal quality at its desired size.
%
\vspace{-0.2cm}
\section{Efficient Supernet RNN-T Compression}
\label{sec:rnnt}
In this section, we describe how a Supernet RNN-T generates subnetwork RNN-Ts of dynamic sizes by varying the width and height of the encoders. It then uses evolutionary search~\cite{yang2022omni} to discover the best encoder architecture of RNN-T at different size constraints.

\vspace{-0.2cm}
\subsection{Constrained Layer and Width Reduction}
\label{sec:layerwidth}
\vspace{-0.15cm}
We experiment with the RNN-T-based ASR models~\cite{graves2013speech}. Supernet training is orthogonal to the type of model we select, and could trivially extend to CTC-based ASR encoders or E2E sequence-to-sequence models.
Our design of the subnetwork search space is pivotal to the success of Supernet training.
Through empirical exploration, we outline three guiding principles for designing the architecture search space for the best Supernet outcomes; the search space should be:
\vspace{-0.1cm}
\begin{enumerate}[wide, labelindent=0.2pt]
\item \textbf{large} enough to find good models of any desired size;\vspace{-0.1cm}
\item \textbf{flexible} enough to discovery new architectures;\vspace{-0.1cm}
\item \textbf{architecture-aware}, focusing on major parameter redundancy in the model.
\end{enumerate}

\begin{figure}[t]
     \centering
     \begin{subfigure}[c]{0.18\textwidth}
         \centering
         \includegraphics[width=\textwidth]{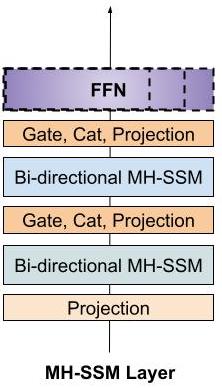}\vspace{-0.2cm}
         \caption{}
         \label{fig:modules}
     \end{subfigure}
     \hfill
     \begin{subfigure}[c]{0.25\textwidth}
         \centering
         \includegraphics[width=\textwidth]{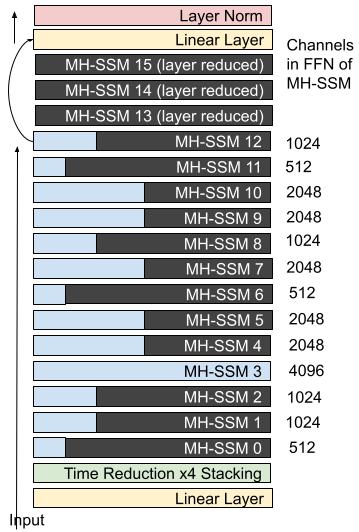}
         \caption{}
         \label{fig:subnetwork}
     \end{subfigure}\vspace{-0.2cm}
      \caption{(a) MH-SSM layer: during Supernet training, subnetworks are created by reducing the output channel sizes of FFN modules (dotted lines) or dropping the entire layer. (b) Encoder channel-and-layer reduction of a 49.6MB Pareto subnetwork for model F. blue: remaining FFN channels, gray: reduced layers and channels.}
        \label{fig:twographs}\vspace{-0.6cm}
\end{figure}

\vspace{-0.1cm}
In RNN-T, the encoder typically occupies the bulk of the parameters. For example, in our baseline model, the encoder occupies 86.9\% of the parameters, followed by the predictor (8.9\%) and joiner (4.2\%).
Therefore, we only use Supernet to find subnetworks of the encoders, while keeping the joiner and decoder unchanged.

Let $\Theta$ be the Supernet model parameters, and $L_n \in \mathbb{Z}_L^n$ denote the list of layers in the RNN-T encoders; the Supernet RNN-T encoder has a maximum of $n$ layers.
We define the width of module $i$, which typically refers to the output channel size of feed-forward networks or linear projection modules, as $ C_i \in {\mathbb{Z}_C}^{m_i} $, 
where this layer has maximum output size of $m_i$. 
A subnetwork can be generated by selecting a subset of layers and a subset of channels from the search space $\bar{\mathbb{Z}}$.
\vspace{-0.1cm}
\begin{equation}
\bar{\mathbb{Z}} \coloneqq \bigcup_0^I \mathbb{Z}_C^{m_i} \cup \mathbb{Z}_L^n
\end{equation}
where $L$ operates on a layer level, and $C$ operates on the sub-layer modular level. To make the search space more manageable, we further constrain the layer and width trimming to a monotonic relationship. That means, if $l_i$ is trimmed, then subsequent $l_{i+1}, ..., l_n$ layers are all removed; similarly, for channels $c_0,...,c_{m_i}$, if the $c_i$ channel is trimmed, all subsequent output channels are also removed.

During training, we perform a sandwich sampling method~\cite{yang2022omni,yu2018slimmable}. We sample four sizes of subnetworks: the maximum (i.e., the entire Supernet), the minimum, and two random sizes (i.e. random selections of $C_i$s and $L_i$s from $\bar{\mathbb{Z}}$). Unlike prior work, we feed to the maximum subnetwork the entire batch of input during the forward step, while to the other three subnetworks a mini-batch that is 1/4 of the batch. This keeps the quality of the maximum network high while keeping the training GPU hours reasonable. We optimize the following loss function:
\begin{equation}
\underset{\Theta}{min} \mathbb{E}_{s \sim \bar{\mathbb{Z}}} \left[
\mathbb{E}_{(\mathbf{x,y^\ast})\sim D_{train}} \mathcal{L}_{rnnt} (\mathbf{y^\ast}|\mathbf{x}; \Theta_s) \right]
\end{equation}
where $s$ is the search space of the Supernet, $\mathcal{L}_{rnnt}(\mathbf{y^\ast} | \mathbf{x}; \Theta_s)$ is the transducer loss~\cite{graves2012sequence} with respect to the correct output sequence $y^\ast$ given input features $x$ from the training data $D_{train}$, computed using subnetwork $\Theta_s$.\vspace{-0.2cm} 

\subsection{Post-training Supernet Pareto Search }
\label{sec:evolutionarysearch}
\vspace{-0.15cm}
We use evolutionary search to discover top-performing subnetworks from a search space of various subnetwork options. The validation dataset WER is used as the fitness score~\cite{yang2023learning}.
The Supernet evolutionary search can dynamically quantize subnetworks, evaluate them on CPUs, and thus consumes $100\times$ less resources than model training. The outcome is a collection of subnetwork configurations, each optimizing accuracy over size constraints $\bar{\mathbb{\tau}} = \left[ \tau_1, \tau_2, ..., \tau_t \right]$:
%
%
%
\begin{equation}
\{ \arg\min_{s_i} \text{\small WER}_{s_i \sim \bar{\mathbb{Z}}, E_{(\mathbf{x,y^\ast}) \sim D_{val}}}(\mathbf{y^\ast}, \mathbf{x}, \Theta_{s_i}), \text{ \small s.t. } \mathbb{M}(s_i) \le \tau_i \}
\end{equation}
where $\mathbb{M}(s_i)$ is the model size of subnetwork $s_i$, and the loss here is word error rate (WER) computed using beam search decoding with beam size=5.

\vspace{-0.2cm}
\section{Improving Supernet Training}\vspace{-0.2cm}
\label{sec:trainsupernet}
In this section, we outline three key strategies for improving the quality of Supernet-trained models. While each technique has been applied in non-Supernet contexts, their combination is novel.
\begin{enumerate}[wide, labelindent=0pt]
\item \textit{Adaptive Dropout:} different subnetworks contribute different magnitudes of gradient L2-norm during Supernet training. Inspired by the adaptive dropout used in Omni-sparsity DNN training~\cite{yang2022omni}, we adjust the magnitude of dropouts in the FFN layers, which immediately follows the modules with reduced output channel during training. Intuitively, a module with reduced dimensionality requires less regularization in order to produce a similar output and gradient norm. 
Therefore, $dropout_{c_i} = dropout_{c_{m_i}} \times \frac{c_i}{m_i}$, immediately follows the reduced output channel size of module $i$, where $m_i$ is the maximum output channel size, and $c_i$ is the current channel size.
\item \textit{Sampled In-place Knowledge Distillation (KD):} during each step of the Supernet training, we use the entire Supernet as the ``teacher'' model, and force in-place distillation of the max-network's output probability distributions onto the sampled subnetwork, i.e. the ``student'', with data from each mini-batch. We experiment with two types of output probability divergence functions: (1) the Kullback–Leibler divergence (KLD), which is known to improve the accuracy of a compressed ASR during knowledge distillation~\cite{pang2018compression,movsner2019improving}; and (2) the Alpha-divergence~\cite{wang2021alphanet} (AlphaD), which has been shown to better capture the teacher network's uncertainties in the output probability distributions than KL in a Supernet training setting. 
Distilling the teacher's output probability distribution over the entire RNN-T lattice is memory-intensive and slows down training by requiring a smaller batch size.
To overcome that, we improve the training memory by subsampling top $j$ output probabilities from the teacher. In a previous study on efficient RNN-T distillation~\cite{panchapagesan2021efficient}, $j$ was set to 2, which meant that the distillation focused on only 3 logit dimensions: the target token, the blank token, and the cumulative sum of the remaining probabilities. We hypothesize that this insufficiently captures the uncertainty of the teacher output probabilities. Therefore, in this work, we compare $j$ values of $[2, 10, 100]$ (distillation probability dimensions = j+1). The loss function is now:
\begin{align}
     \mathcal{L} &=  \lambda \mathcal{L}_{KD} + \mathcal{L}_{rnnt}\label{eq:kld} \\
   \underset{s \in \bar{\mathbb{Z}}} {\mathcal{L}_{KD}(\Theta_s; \Theta)} &= \mathbb{E}_{x\in D_{mini-batch}} \left[ f(p_j(x;\Theta) || q_j(x; \Theta_s)  \right] 
\end{align}
where we use default $\lambda = 1.0$, $f$ is KLD or AlphaD, computed over the sampled probability of top $j$ tokens, and the sum of the rest ($p_j$ being the teacher's distributions, $q_j$ the student's). For Alpha-divergence, we use the default setting of $\alpha_{-}$=-1, $\alpha_{+}$=1 and $\beta$=5.0. 
\item\textit{ScaledAdam Optimizer:} we explore training the Supernet with the ScaledAdam optimizer~\cite{yao2023zipformer}. As a variant of Adam optimizer, the ScaledAdam scales each parameter's update based on its norm. Intuitively, the ScaledAdam optimizer improves gradient stability of the Supernet training, where each subnetwork contributes drastically different gradient norms to model parameters during mini-batch forward-backward step.
\end{enumerate}

\vspace{-0.2cm}
\section{Experiment Setup}\vspace{-0.2cm}
\label{sec:exp}

\begin{figure}[t!]
\centering
\includegraphics[width=8.0cm,height=4.2cm]{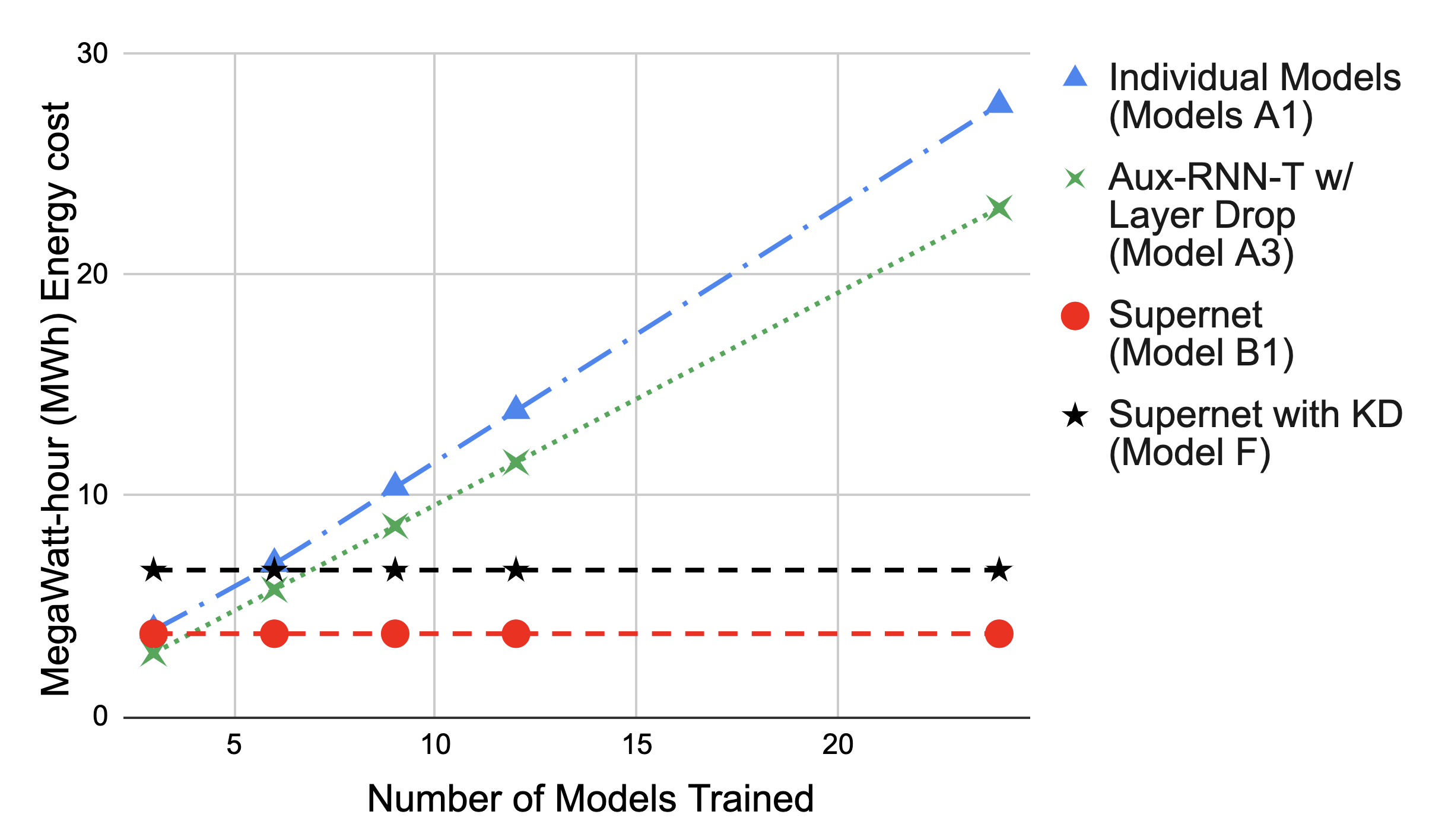}\vspace{-0.2cm}
\caption{Energy (MWh) consumed by training models at scale.}\label{fig:energy}\vspace{-0.5cm}
\end{figure}

In this paper, we validate the results of TODM Supernet on a non-streaming Multi-Head State Space Model (MH-SSM) RNN-T. The encoder of our RNN-T is built with 16 MH-SSM layers; these layers are attention free and have been demonstrated to achieve on-par performances compared to transformer-based ASR~\cite{fathullah2023multi}. Each MH-SSM layer consists of 2-stacked, 1-headed MH-SSM, and 4096 feed-forward net (FFN) dimension. Our RNN-T architecture follows the example set in prior work~\cite{fathullah2023multi}: three 512-dimensional LSTM layers plus a linear layer in the predictor; a linear project layer and a ReLU gate in the joiner. 
In total, our non-streaming MH-SSM RNN-T has 99.9 million parameters. During Supernet training, we selectively reduce entire layers of MH-SSM, or alter the size of the FFN module to $C_i \in [512, 1024, 2048, 4096]$. The FFN consists of 77.8\% of the total number of parameters in a MH-SSM layer, and has high potential of parameter redundancy. Layer reduction in MH-SSM supernet is performed as reducing top 0, 3, or 7 MH-SSM layers. See Fig~\ref{fig:modules} for illustration.
All our baseline MH-SSM RNN-T ($A1, A2, A3$ in Table~\ref{tab:ablation}) are trained with the same hyper-parameter set up: a total of 180 epochs of training, under a fixed 0.006 learning rate, force-anneal at 60 epochs with shrinking factor 0.96; 0.1 weight decay; Adam optimizer with $\beta_1=0.9, \beta_2=0.999$. 

We train the RNN-Ts with the LibriSpeech 960 hr training data~\cite{librispeech}. We obtain 80-dimensional log Mel-filterbank features from each 25 ms audio window, sliding the window ahead every 10 ms. We use a pre-trained 4096-dimensional sentence piece vocab~\cite{kudo2018sentencepiece}, plus a `blank' symbol, as the RNN-T target. After Supernet training, we use the 10.7h LibriSpeech dev-clean and dev-other data to identify best subnetwork architectures during evolutionary search.

All model training is done with 32 NVIDIA-A100 GPUs; to prevent GPU queue preemption, CPU scheduling wait-times, and other data center down-times from polluting the GPU-hour computation, we report the total energy expenses of each model training in terms of MegaWatt-hour~(MWh), measured by the energy consumed and recorded by our GPU cluster.

\vspace{-0.2cm}
\section{Results and Discussion}
\label{sec:results}
\vspace{-0.2cm}

We compare our TODM Supernet (Model F) with 4 baselines:

\begin{figure}[t!]
\centering
\includegraphics[width=8.2cm,height=4.4cm]
{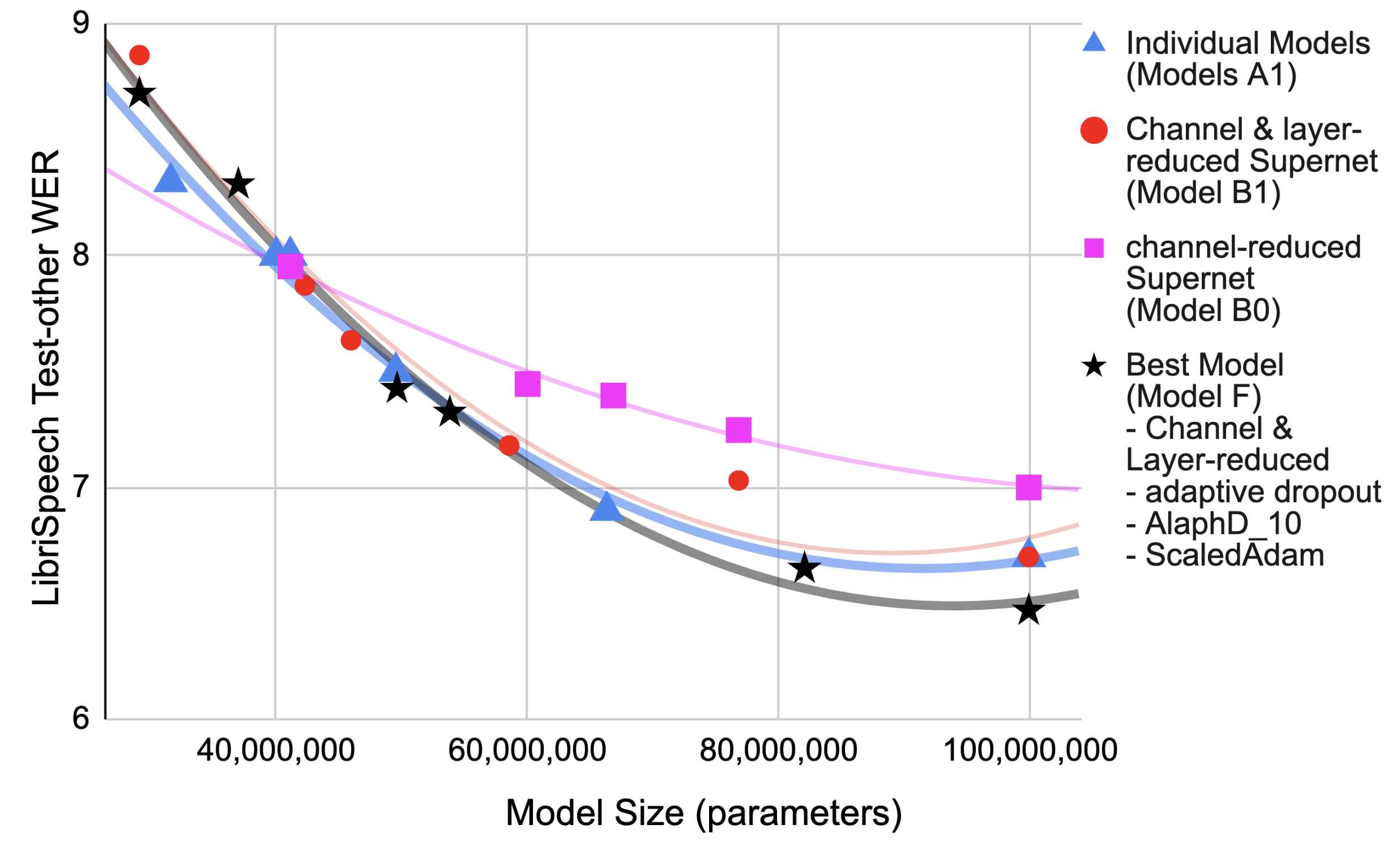}\vspace{-0.2cm}
\caption{LibriSpeech test-other WER vs. model sizes. Our proposed method (Model F) produces subnetworks that are on-par with, or exceeds best individually trained models (especially for models $\geq$ 50 million parameters). The trend lines are second degree polynomial approximations of the results.}
\label{fig:layerdrop}\vspace{-0.3cm}
\end{figure}

\vspace{-0.1cm}
\begin{enumerate}[wide, labelindent=0.1pt]
    \item \textit{A1} represents a collection of individually trained and tuned models, the sizes of which are determined by reducing layer sizes, channel sizes, or both, from the largest MH-SSM model. The energy used in training scales proportionally to the number and sizes of the models, we are thus capable to explore only limited architectures.\vspace{-0.1cm}
    \item \textit{A2} represents the results of the largest model, trained with auxiliary cross-entropy loss~\cite{auxloss} at layers 8, 12, and 16; auxiliary training introduced 1.15 million more parameters to the model, but did not end up improving the model performances in Table~\ref{tab:ablation}. \vspace{-0.1cm}
    \item \textit{A3} is the results of the largest model, trained with auxiliary RNN-T loss at layers 8, 12, and 16 (i.e. layer-drop) -- each of these layers is accompanied with a unique combination of linear layer, gates and layer-norm, which is then fed into a shared predictor and joiner~\cite{auxloss}. For $k$ subnetworks in the auxiliary RNN-T loss training, there is $k\times$ resources increase during training, and an additional model size of $k\times$0.53 million parameters. Note that A2 and A3 exhibit comparable training energy consumption. We use them to investigate the extra regularization effects when loss functions are applied to the intermediate layers of the RNN-T model. \vspace{-0.1cm}
    \item \textit{B0} represents a channel-reduction-only Supernet. It is a generalized version of the slimmable network~\cite{yu2018slimmable}, in which subnetworks of different layer-wise widths co-exist in a Supernet. \vspace{-0.1cm}
\end{enumerate} \vspace{-0.1cm}

\vspace{-0.3cm}
\subsection{On the Energy Consumed during Training} \vspace{-0.2cm}
In Fig~\ref{fig:energy} we show the energy consumed by training up to $X\in[3,6,..30]$ number of models. The energy cost of training each Supernet is constant; the energy costs scale linearly with the number of non-Supernet models (Models A1 and A3). Supernet with KD (Model F) consumes more energy than Supernet without KD (Model B), due to the need of using smaller batches sizes and thus longer training time, as the memory consumption of storing gradients and logits for KD is non-trivial. 

\begin{table}[t] 
\begin{center}\setlength{\tabcolsep}{0.18em}
\begin{tabular}{ |c|l||c|c||c|c||c|c|} 
 \hline
 Model & Baselines &WER& size&WER& size & WER & Size\\\hline
 A1 & 3 models & 6.7 & 99.9 & \textbf{8.3}& 31.6 & 7.5 & 49.5 \\\hline
 A2 & auxCE$_{8,12,16}$& 6.7 & 101.1  \\\hline 
 A3 & auxRNNT$_{8,12,16}$&6.8 & 101.5 & 8.8 & 40.0 & 7.2 & 53.1  \\\hline\hline
 Model & Supernet Pareto& WER & size & WER & size & WER & size \\\hline
 B0 & channel only reduction& 7.0 & 99.9 & 8.0 & 41.1 & 7.4 & 60.0 \\\hline
 B1 & layer-channel reduction& 6.7 & 99.9 & 8.9 & 29.1 & 7.6 & 45.9 \\\hline
 C & B1+adaptive dropout & 6.8 & 99.9 & \textbf{8.7} & 29.1 & 7.5 & 47.0 \\\hline 
 D & C+KLD$_{10}$ & 6.7 & 99.9 & 9.1 & 29.1 & 7.7 & 49.1 \\\hline
 E & C+alphaD$_{10}$ & 6.7 & 99.9 & 9.2 & 29.1 & 7.8 & 49.6 \\\hline
 F &E$_{120epoch}$ ScaledAdam& \textbf{6.5} & 99.9 & \textbf{8.7} & 29.1 & 7.4 & 49.6 \\\hline
\end{tabular}\vspace{-0.2cm}
\caption{Word Error Rates of models on the LibriSpeech test-other data set. We show WERs of 3 best model sizes from each training job if possible: $\sim$99, $\sim$30, and $\sim$50 MB, obtained along the Pareto front via evolutionary search. Model sizes are determined after 8-bit quantization. Results are also illustrated in Fig~\ref{fig:layerdrop}.}
\vspace{-0.2cm}
\label{tab:ablation}
\end{center}\vspace{-0.6cm}
\end{table}

\vspace{-0.2cm}
\subsection{On the Design of Supernet Search Space}
\label{sec:slimmable}
\vspace{-0.2cm}
We lay out three principles for the Supernet search space design in Section~\ref{sec:layerwidth}. We demonstrate these using two Supernets in Fig~\ref{fig:layerdrop}. The square magenta dots (Model B0) show Pareto results of the channel-reduction Supernet, otherwise known as the slimmable networks~\cite{yu2018slimmable}; the round red dots (Model B1) are Pareto results trained with layer-and-channel reduction. Layer reduction expands the search space, allowing the Supernet Pareto search to discover more effective medium sized models, hence resulting in better model accuracy vs model size trade-offs.

\vspace{-0.2cm}
\subsection{On Probability Sampling for KD}
\vspace{-0.1cm}
To examine the effectiveness of KD, we evaluated Supernet results at 120 epochs, 2/3 of the total training time. We record the WERs of the max (i.e. entire Supernet), the min, and a fixed architecture (47MB) models in Table~\ref{tab:distillation}. The fixed architecture does not always lie on the Pareto front of each Supernet.
Due to GPU out-of-memory failures that occur during training with KD of the full 4096-sized logits; ``C+KLD$_{4096}$" is thus not present in Table~\ref{tab:distillation}.
We find that KD helps the supernet converge faster. The model G$_{120 epoch}$ with KD already exceeds the baseline model C$_{120 epoch}$'s WER by relative 10.3\%, 4.2\%, and 2.4\% in its max, min, and fixed-size networks.

Surprisingly, with KD, the max network converges much faster than the smaller networks, even though KD is designed to improve the representations of the smaller networks. We hypothesize that it is because high-quality subnetworks contribute to the overall max network's performances via weight sharing. 

AlphaD and KLD produce similar results, with AlphaD slightly outperforming KLD by 2-3\%. However, the difference is not statistically significant. 
AlphaD could prevent overestimation and underestimation of the teacher's uncertainties~\cite{wang2021alphanet}, but sampling logits and reducing the probability distribution may already underestimate the teacher's uncertainty, limiting the effectiveness of AlphaD.

\begin{table}[t]\vspace{-0.2cm}
\begin{center}\setlength{\tabcolsep}{0.2em}
\begin{tabular}{|l|l|c|c|c|}
\hline
 Model & KD$_{sample prob}$ & WER$_{99.9MB}$& WER$_{29.1MB}$ &  WER$_{47.0MB}$ \\\hline
 C$_{120epoch}$ & NO KD & 7.8 &  9.6 & 8.4 \\\hline\hline
 G$_{120epoch}$ & C+alphaD$_{2}$ & 7.0   & \textbf{9.2} &  8.2   \\\hline
 E$_{120epoch}$ & C+alphaD$_{10}$ & \textbf{6.9} & 9.4  & 8.1 \\\hline
 I$_{120epoch}$ & C+alphaD$_{100}$ &7.0 & 9.5 & 8.1 \\\hline\hline
 J$_{120epoch}$ & C+KLD$_2$ & 7.2 & 9.4 & 8.1 \\\hline
 D$_{120epoch}$ & C+KLD$_{11}$ & 7.1 & 9.4  & \textbf{8.0} \\\hline
 L$_{120epoch}$ & C+KLD$_{100}$ & 7.2 & 9.6 & 8.2\\\hline
\end{tabular}\vspace{-0.2cm}
\caption{Supernet trained with probability sampling during in-place knowledge distillation(KD), with alpha divergence (alphaD) or KL divergence(KLD). Subscripts denote sampling range j$\in$[2,11,100]. Results are evaluated at 120 epochs for max(99.9MB), min(29.1MB) and a fixed architecture (47.0MB) on LibriSpeech test-other dataset.}
\label{tab:distillation}\vspace{-0.8cm}
\end{center}
\end{table}

\vspace{-0.2cm}
\subsection{On the Ablation of Training Strategies}
\vspace{-0.1cm}


In the Supernet results in Table~\ref{tab:ablation},
the last two columns of the WER-Size pairs are results of the best model on the Pareto front when searching for a RNN-T subnetwork $\sim$50 million parameters. Model B1 is trained with channel-and-layer reduction; Model C adds adaptive dropout during training; Model D uses both adaptive dropout and KL-divergence with sampled top 10 log-probabilities; Model E, similar to Model D, uses Alpha-divergence instead; Model F uses adaptive dropout, Alpha-divergence (sampled over 10 log-probabilities), and ScaledAdam optimizer fine-tuning after 120 epochs. All models are trained from scratch and evaluated at 180 epochs. 


Model A3 in Table~\ref{tab:ablation}, trained with RNN-T losses on intermediate layers, is a special case of Supernet B1 -- weight sharing between 3 layer-dropped subnetworks. Model A3 has, however, significantly worse performance at 40MB compared to Model B1. 
Auxiliary regularization and layer-dropping alone don't explain the Supernet's superior performance compared to A2 and A3. 

Comparing Models B1 and C, adaptive dropout alone improves the Supernet training by only a small margin. Despite the early convergence of D$_{120 epoch}$ and E$_{120 epoch}$, models D and E subsequently converge much slower than model C. We observe that while in-place KD helps the Supernet training to converge at earlier stages of the training, it also causes the validation loss to diverge at around 100 to 120 epochs. We thus switch to using the ScaledAdam optimizer at 120 epochs, and reduce the KD loss weight in Equation~\eqref{eq:kld} to $\lambda=0.1$. Even though the ScaledAdam optimizer is helpful for the last 60 epochs of training, we observe that training TODM from the very beginning with it prevents the Supernet from converging fully.

\vspace{-0.4cm}
\subsection{Other Observations}
\vspace{-0.2cm}
We find that Supernet models do not benefit from longer training time. In fact, continuing training Supernet Model C for another 60 epochs results in a relative WER rise by 16.4\% and 8.0\% for the max and min models respectively. This suggests that TODM converges in a similar number of epochs as training one of the models alone. 
%
We also find that increasing the learning rate or seeding the Supernet with a pre-trained model hurts Pareto subnetwork accuracies. The latter indicates that the best weight-sharing Supernet may differ significantly from the max-network if it is trained alone.

Finally, the Pareto front model architecture of Model F at 49.6MB in Fig~\ref{fig:subnetwork} does not have obvious architectural patterns. This may explain why it is difficult to arrive at this architecture through manual tuning. 

Throughout our experiments, the trends in WER for test-clean dataset are similar to those for test-other -- for example, Models in A1 have test-clean WERs of 2.6 (99.9MB) and 3.5 (31.6MB); Supernet Model F has WERs 2.5 (99.9MB) and 3.4 (29.1MB). 

\vspace{-0.2cm}
\section{Conclusion}
\label{sec:con}
\vspace{-0.2cm}
This paper presents the TODM framework, which can train and discover optimized, dense RNN-Ts of various sizes from a single Supernet, with comparable training resources to a single-model training job. 
We introduce three strategies to TODM to improve Supernet outcomes: adaptive dropout, in-place sampled knowledge distillation, and ScaledAdam optimizer fine-tuning. TODM discovers many models along the Pareto front of accuracy vs. size, which would have be resource-intensive to find and train manually via trial-and-error.

\bibliographystyle{IEEEbib}
\bibliography{refs}

\end{document}